\documentclass[runningheads]{llncs}
\usepackage{cite}
\usepackage{graphicx}
\usepackage{multirow}
\usepackage{amssymb}
\usepackage{array}
\usepackage{url}
\usepackage{enumerate}
\usepackage{enumitem}
\usepackage{amsmath}
\usepackage{booktabs}
\usepackage{multirow}

\usepackage{algorithm}
\usepackage{algpseudocode}
\usepackage{bm}

\begin{document}

\title{Less is More: Adaptive Curriculum Learning for Thyroid Nodule Diagnosis}

\author{Haifan Gong$^{1, 3, \dag}$, Hui Cheng$^{1, \dag}$, Yifan Xie$^1$, Shuangyi Tan$^{3,4}$, Guanqi Chen$^{1}$,\\ Fei Chen$^{2\star}$, Guanbin Li$^{1}$\thanks{Corresponding authors are Fei Chen~(gzchenfei@126.com) and Guanbin Li~(liguanbin@mail.sysu.edu.cn). $\dag$Haifan Gong and Hui Cheng contribute equally to this work.}}

%
\institute{$^1$ School of Computer Science and Engineering, Sun Yat-sen University, Guangdong, China 
\\ $^2$ Zhujiang Hospital, Southern Medical University, Guangdong, China \\ $^3$ Shenzhen Research Institute of Big Data, Shenzhen, China \\ $^4$ The Chinese University of Hong Kong, Shenzhen, China}

\authorrunning{H. Gong et al.}

\maketitle


\begin{abstract}
Thyroid nodule classification aims at determining whether the nodule is benign or malignant based on a given ultrasound image. However, the label obtained by the cytological biopsy which is the golden standard in clinical medicine is not always consistent with the ultrasound imaging TI-RADS criteria. The information difference between the two causes the existing deep learning-based classification methods to be indecisive. To solve the {\em Inconsistent Label} problem, we propose an Adaptive Curriculum Learning (ACL) framework, which adaptively discovers and discards the samples with inconsistent labels. Specifically, ACL takes both hard sample and model certainty into account, and could accurately determine the threshold to distinguish the samples with  {\em Inconsistent Label}. Moreover, we contribute TNCD: a Thyroid Nodule Classification Dataset to facilitate future related research on the thyroid nodules. Extensive experimental results on TNCD based on three different backbone networks not only demonstrate the superiority of our method but also prove that the less-is-more principle which strategically discards the samples with {\em Inconsistent Label} could yield performance gains. Source code and data are available at https://github.com/chenghui-666/ACL/.

\end{abstract}

\section{Introduction}
Thyroid nodule is a common clinical disease with an incidence of 19\%-68\% in the population, where about 5\%-15\% of them are malignant \cite{chen20review}. 
Ultrasonic image-based diagnosis is the most widely used technique to determine whether the thyroid nodule is benign or malignant because of its low cost, efficiency, and sensitivity. 
However, unlike standardized CT \& MRI images, ultrasonic images are taken at variant positions from different angles. Meanwhile, ultrasonic images are susceptible to noise due to their low contrast, which is challenging for inexperienced radiologists to perform diagnoses \cite{gong2021multi}. Thus, it is valuable to design an accurate computer-aided diagnosis (CAD) system to reduce the performance gap between inexperienced radiologists and experienced ones.

To provide guidance for the diagnosis of thyroid nodule, Tessler et al. \cite{tessler2017tirads} have proposed the Thyroid Imaging Reporting And Data System (TI-RADS) which describes five groups of appearance feature-based diagnostic criteria to conduct qualitative analyses for thyroid nodules.
Nevertheless, in clinical medicine, the fine needle aspiration (FNA) \cite{paschke17fna} based cytological biopsy is the golden standard for the diagnosis of the thyroid nodule. 
In fact, the results of the diagnostic analysis guided by TI-RADS are not necessarily consistent with the judgment based on the golden rule of pathology. TI-RADS, as a diagnostic manual based on the empirical summary, judges the nature of nodules through image features, which means that benign nodules are bound to have some characteristics of malignancy, and malignant nodules must also contain benign morphological features. In terms of probability, the label itself inevitably has uncertainty.

Therefore, the label obtained by the cytological biopsy is not always consistent with that from TI-RADS, and this causes an {\em Inconsistent Label} problem: the samples with inconsistent labels tend to be harder for the model to fit, which leads to the complexity of the decision plane and harms the generalization of the model. Based on the above concerns, we propose an adaptive curriculum learning framework to resolve these issues. The contribution of this work can be summarized as follow: (1) We propose a curriculum learning-based algorithm to resolve the inconsistent label problem. Specifically, it works by adaptively discovering and discarding the hard samples; (2) We contribute TNCD: a benchmark for the thyroid nodule classification task, to further encourage the research development of thyroid nodule diagnosis. (3) Extensive experiments compared with other state-of-the-art methods have demonstrated the effectiveness of our method, and proving the less-is-more proverbial that end-to-end learning with fewer samples could achieve better performance.

\section{Related Work}
\subsubsection{Thyroid Nodule Diagnosis}
Deep neural networks (DNN) have shown their dominance in the field of image representation learning. Based on DNN, Wang et al. \cite{wangAutomatic20} proposed an attention-based network that aggregates the extracted features from multiple ultrasound images. 
Song et al. \cite{songthyroid20} proposed a hybrid feature cropping network to extract discriminative features for better performance on classification, and this network reduces the negative impacts of local similarities between benign and malignant nodules. Zhao et al. \cite{zhao2022local} proposed a local and global feature disentangled network to segment and classify the thyroid nodules. To make the automatic diagnosis more accurate and consistent with human cognition, several works \cite{liuautomated19, yangintergrate21,avola2021multimodal} were proposed to integrate domain knowledge for thyroid nodule diagnosis, such as boundary feature, aspect ratio, echo pattern, orientation, etc.
However, all the previous works ignore the {\em Inconsistent Label} issue. To address this issue, we proposed a curriculum learning-based algorithm to discard the samples adaptively during the neural network training stage.

\subsubsection{Curriculum Learning}
Curriculum learning \cite{Bengio09CL} is a training strategy that makes models to gradually learn from easy to hard. Many works adopt curriculum learning to improve the generalization or convergence speed of models in the domain of computer vision \cite{WangGYWY19} or natural language processing \cite{PlataniosSNPM19}. 
Recently, curriculum learning-based approaches \cite{Castells20superloss, LyuT20curriloss} have been used to overcome the inherent hardness of learning from hard samples, as it automatically decreases the weight of samples based on their difficulty. Lyu et al. \cite{LyuT20curriloss} proposed to adaptively select samples with a tighter upper bound loss against label corruption. Castells et al. \cite{Castells20superloss} designed SuperLoss that mathematically decreases the contribution of samples with a large loss. Liu et al. \cite{liu2021co} proposed a co-correcting framework to relabel the images with dual-network curriculum learning. However, all the above-mentioned methods neither rely on the hard-craft schedule nor ignore the samples with inconsistent labels. Thus, we develop an adaptive curriculum learning paradigm to adaptively discover and discard these samples.

\begin{figure*}[tbp]
\includegraphics[width=\textwidth]{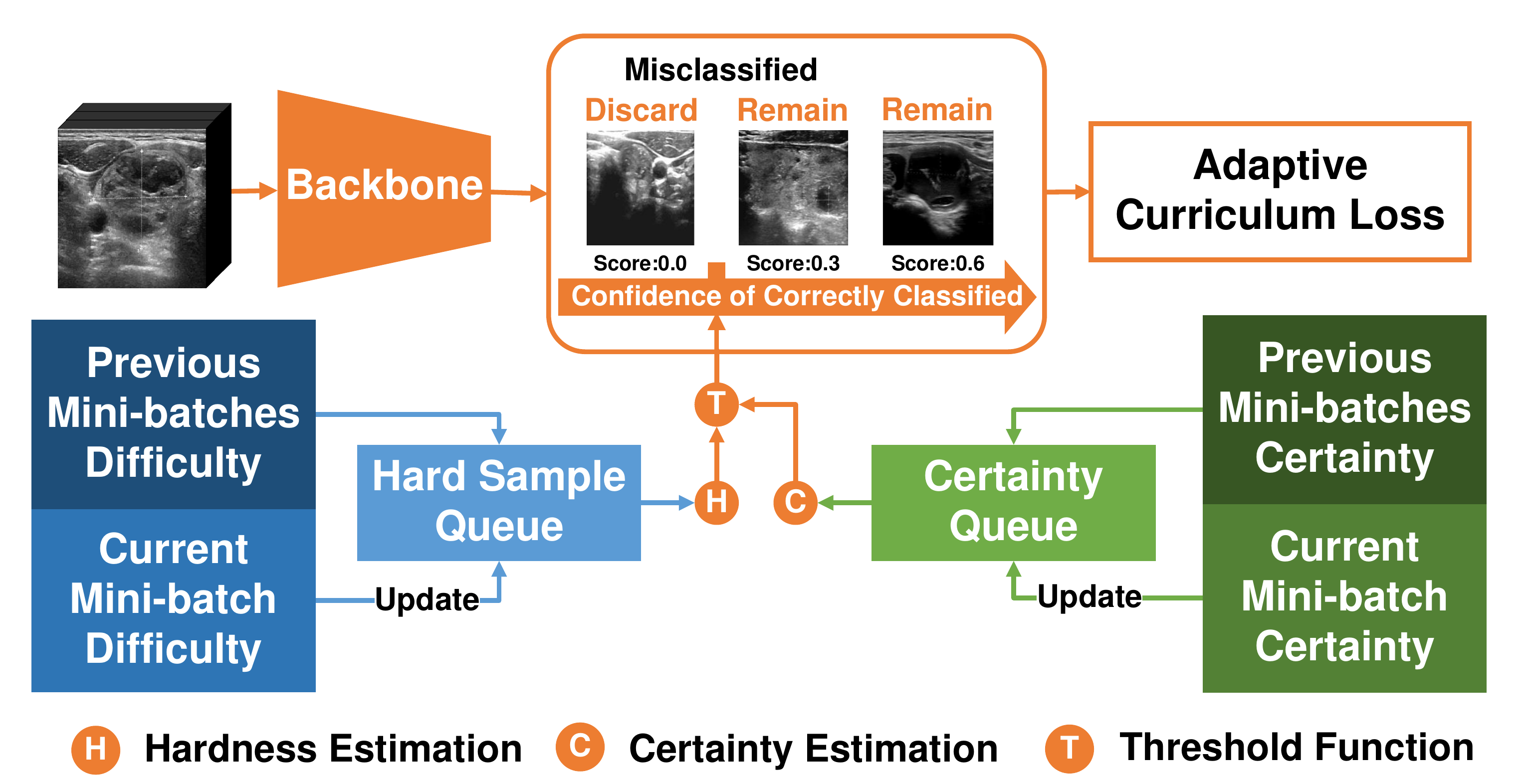}
\caption{Overview of the proposed adaptive curriculum learning framework. The orange
part shows the training pipeline on a mini-batch. The hard sample estimation queue is
shown in blue blocks, while the certainty estimation module is shown in green blocks.} \label{fig1}
\end{figure*}

\section{Methodology}
We propose an adaptive curriculum learning (ACL) framework for thyroid nodule diagnosis, which is shown in Fig.\ref{fig1}. ACL is mainly composed of two parts: a curriculum learning-based sample scheduler and a model certainty estimating function. The idea underlying ACL is to mitigate the {\em Inconsistent Label} problem by adaptively discovering and discarding the detrimental samples, thus forcing the model to adaptively learn from easy samples to hard samples.


\subsection{Hard Sample Discovery with Confidence Queue}
The core idea of curriculum learning is to gradually excavate hard samples and adjust their weights adaptively in the process of model training. Therefore, we need to design a function to discriminate the hardness of a sample. Considering that the samples with inconsistent labels tend to be misjudged with high confidence, we first define the set $P_c$ that contains the sample's confidence $c_i$ as:

\begin{equation}
    P_c = \{c_0, ... ,c_i\}
\end{equation}
where $c_i = P(x_i|y_i)$. $x_i$ and $y_i$ denote the predicted probability (after Softmax function) and its corresponding label, respectively. $i$ is the index of the sample. $c_i$ ranges from 0 to 1. When $c_i < 0.5$ the model makes incorrect prediction. Thus, the set $H$ of hard samples in current mini-batch $c'_i$ is defined as:
\begin{equation}
    H = \{c_i \in P_c | \mathbb{I}(y_i \neq y'_i) \}
\end{equation}
where $y'_i$ is the predicted label. $\mathbb{I}(\cdot)$ is an indicator function that returns true when condition meets, which means the correctly classified $i^{th}$ sample will not enter the hard sample queue. Therefore, the hard sample set $H$ will be fed into the queue to calculate the threshold, which is used to distinguish the samples with the inconsistent label.

To find the hard samples more accurately, we propose the adaptive thresholding function $T_{ada}$. The tailor-designed  $T_{ada}$ can adaptively constrain the model to better learn the easy samples in the early stage, and gradually learn the hard samples in the later stage.
As a dynamic threshold, $T_{ada}$ should be able to (1) update in time; (2) estimate with sufficient samples.
Thus, we design a hard sample confidence queue $Q_h$ with fixed length $L$ to store the confidence of misclassified samples across batches.
This queue is updated every batch by following the first in first out (FIFO) rule, which means we first dequeue the oldest $k$ elements in $Q_h$, then send the set $H$ into $Q_h$.
Let the queue before updating be $Q^{pre}_h\{c_1, ..., c_l\}$ where the subscripts denote the previous position of the elements, this process is formulated as:
\begin{equation}
    Q_{h} = 
    \begin{cases}
        \{c_1, ..., c_l, ..., c_{l+k} \}, & l \leq L - k \\
        \{c_{l+k-L}, ..., c_l, ..., c_{l+k} \}, & L-k < l < L \\
        \{c_{k}, ..., c_l, ..., c_{l+k} \}, & l = L \\
    \end{cases}\\,
\end{equation}
where $L$ and $l$ denotes the fixed length and previous length of the queue, respectively. The number of misclassified sample is represented by $k$. Let $\mu$ and $\sigma$ be the average value and standard deviation of elements in $Q_h$, the formulation of $T_{ada}$ is defined as:
\begin{equation}
    T_{ada} = \mu + \alpha \cdot \sigma
\end{equation}
where $\alpha$ is a hyper-parameter to trade-off the numbers of the hard samples. 


\subsection{Model Certainty Estimation with Certainty Queue}
Considering that the $T_{ada}$ did not take the model's certainty into concern, which will lead to the inaccurate estimation of $T_{ada}$, we propose to embedded the model's certainty into the $T_{ada}$ by replacing the hyper-parameter $\alpha$ with a \textbf{self-configured \& certainty-aware} variable $\mathbb{\theta}$. The idea underlying this modification is to use the model's certainty to constrain the standard deviation of difficult sample queues to better estimate difficult samples. As the model certainty can be reflected by the model's prediction, we define the sample certainty $\hat{c}_i$ as:
\begin{equation}
    \hat{c}_i = max(x_i)
\end{equation}
where $x_i$ is the model's prediction after Softmax function. The value of the sample $\hat{c}_i$ ranges from (0.5, 1). The larger value is, the more certain the model is. After that, we follow the idea of a hard sample queue in Section 3.1 and propose a certainty queue $Q_c$ to estimate the model's certainty. The difference between $Q_c$ and $Q_h$ is that $Q_c$ uses all the samples in the current mini-batch for updating, while $Q_h$ only updates the queue with the misclassified samples. Since queue length $L$ is set equal to the length of $Q_h$ which is a multiple of batch-size $B$, letting the queue before updating be $Q^{pre}_c\{\hat{c}_1, ..., \hat{c}_l\}$, this process could be formulated as:
\begin{equation}
    Q_c = 
    \begin{cases}
        \{\hat{c}_1, ..., \hat{c}_l, ..., \hat{c}_{l+B} \}, & l \leq L - B \\
        \{\hat{c}_B, ..., \hat{c}_l, ..., \hat{c}_{l+B} \}, & l = L \\
    \end{cases}\\
\end{equation}
Thus $\mathbb{\theta}$ is calculated by averaging the elements in queue $Q_c$ with $\theta = \frac{1}{L} \sum^{L}_{j = 1} Q^c_j$.
Intuitively, the $\theta$ should gradually change from 0.5 to 1 as the model's iteration. In the early stage of training, the model is uncertain about the samples, we should discard fewer samples to force the model to learn more imaging features. In the late stage of training, the model is sure about the samples. Thus, we strategically multiply $\theta$ and standard deviation $\sigma$ to embed the model certainty into hard sample selection, forcing the $T_{ada}$ to better distinguish the hard samples. Thus, Equation. 4 is re-written as:
\begin{equation}
    T_{ada} = \mu + \theta \cdot \sigma
\end{equation}
The overall pseudo code to update the $T_{ada}$ is shown in Algorithm \ref{alg:Tada}.


\begin{algorithm}
\caption{Updating $T_{ada}$ in $b^{th}$ batch}\label{alg:Tada}
\begin{algorithmic}
\Require $B \in N^+$, the batch size

\ \ \ \ \ $X \in \mathbb{R}^{B \times 2}$, the prediction all samples in $b^{th}$ batch

\ \ \ \ \ $Y \in \{0,1\}^{B}$, the labels of all samples in $b^{th}$ batch

\ \ \ \ \ $Q_h \in (0, 0.5)^{N}$, the hard sample queue updated after $(b-1)^{th}$ batch

\ \ \ \ \ $Q_c \in (0.5, 1)^{N}$, the certainty queue updated after $(b-1)^{th}$ batch

\Ensure The updated $T_{ada}$
\State $P = softmax(X)$  \Comment{The predicted possibility $ P \in \mathbb{R}^{B \times 2}$}
\State $Y' = argmax(P[:])$  \Comment{The predicted label $Y' \in \{0,1\}^{B}$}
\For{i \textbf{in} range(0,B)}
    \State $C = max(P)$ \Comment{The set of certainties.}
    \State $dequeue(Q_c, len(C))$ \Comment{Dequeue the earliest $len(C)$ elements}
    \State $enqueue(Q_c, C)$ \Comment{Enqueue the probability of the predicted label}
    \If{$Y'[i] \neq Y[i]$}
        \State $H = \{P_i\in P|Y'[i] \neq Y[i]\}$  \Comment{The set of difficulty.}
        \State $dequeue(Q_h, len(H))$ \Comment{Dequeue the earliest element}
        \State $enqueue(Q_h,H)$ \Comment{Enqueue the confidence of misclassified sample}
    \EndIf
\EndFor
\State $\theta = mean(Q_c)$ \Comment{Estimate the model certainty}
\State $\mu = mean(Q_h)$
\State $\sigma = std(Q_h)$
\State $T_{ada} = \mu + \theta \cdot \sigma$ \Comment{Update $T_{ada}$ with queue $C$ and model certainty}
\State \Return $T_{ada}$
\end{algorithmic}
\end{algorithm}

\subsection{Loss Function of Adaptive Curriculum Learning}


With the $T_{ada}$ and the confidence $c_i$ defined in Equation. 1, the curriculum learning process can be described by adjusting the loss $l_{i}$ of $i$-th sample. Let $CE$ be the cross-entropy loss, following the "less  is more" proverb, this process is defined as:
\begin{equation}
    l_{i} = \begin{cases}
               CE(y'_{i}, y_i), & c_{i} \geq T_{ada} \\
               0, & c_{i}<T_{ada} \\
            \end{cases}\\,
\end{equation}
Finally, let $B$ be the batch size, the overall loss $L$ is defined as:
\begin{equation}
    L = \sum^{B}_{i=1}{l_{i}}.
\end{equation}

\section{TNCD: Benchmark for Thyroid Nodule Classification}
To facilitate future research in thyroid nodule diagnosis, we contribute a new Thyroid Nodule Classification Dataset called TNCD, which contains 3493 ultrasound images taken from 2421 patients. According to the results of nodular cytological biopsies, each image is labeled as benign or malignant according to its pathological biopsy result. To verify the performance of the algorithms, the TNCD dataset is divided into the training set and test set by ensuring the images from the same patient only appear in a certain subset. The training set contains 2879 images with 1905 benign and 974 malignant images, while the test set contains 614 images with 378 benign and 236 malignant images.

\section{Experiment}
\subsection{Implementation and Evaluation Metric}
The framework is implemented in PyTorch 1.11. All models are trained with NVIDIA 3090 GPU with CUDA 11.3, and initialized by the ImageNet pre-trained weights. Stochastic gradient descent is used to optimize our models at an initial learning rate of 0.001, `Poly' learning rate policy is applied, where $lr=lr_{init}\times(1-\frac{epoch}{epoch_{total}})^{{0.9}}$. Batch size and training epoch are set to 16 and 50, respectively. All images are resized to $224 \times 224$ with horizontal flip argumentation. We performed oversampling on the malignant samples to avoid the over-fitting of the majority benign samples. It is worth noting that as the training process is not stable initially, we trained the model with the cross entropy loss for the first three epochs, then trained the model with the proposed ACL loss.
We use accuracy, precision, recall, F1-score and AUC as metrics.

\begin{table}[]
\centering
\caption{Ablation study of the queue length and the hyper-parameter.}
\begin{tabular}{@{}ccccc@{}}
\toprule \toprule
Queue length $L$ &   0 (Baseline) &  16  &  32  &  64        \\ \midrule
AUC   &  $77.31_{\pm1.04}$  &  $79.24 _{\pm0.98}$ & $\bm{79.27 _{\pm1.26}}$           &      $79.18 _{\pm1.17}$          \\ \midrule \midrule
Hyper-parameter $\alpha $ &  0  &     1     & 2     & \textbf{$\theta$(Ours)}  \\ \midrule
AUC       &  $79.27 _{\pm1.26}$&   $79.66 _{\pm1.13}$    &   $79.41 _{\pm1.26}$& $\bm{79.89 _{\pm0.89}}$\\ \bottomrule  \bottomrule
\end{tabular}
\label{abla}
\end{table}

\begin{figure}[tbp]
\includegraphics[width=\textwidth]{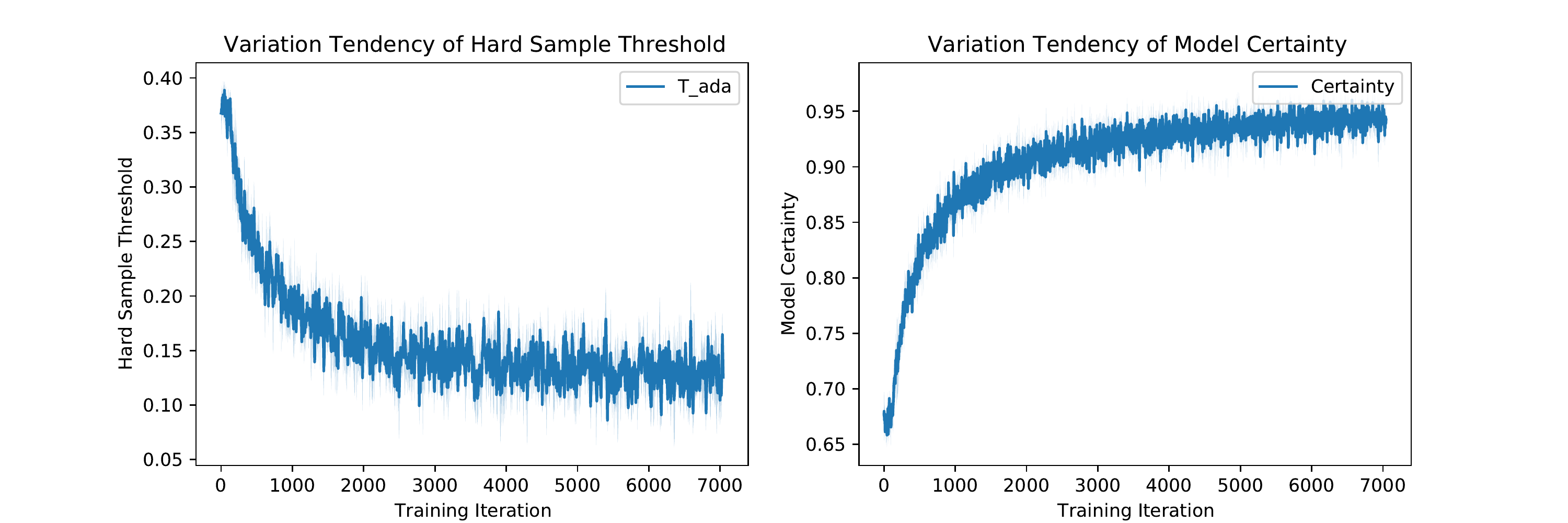}
\caption{The visualization result of the $T_{ada}$ and the model certainty $\theta$.}
\label{fig2}
\end{figure}

\subsection{Ablation Study and Schedule Analysis}
All the ablation studies are conducted based on the ResNet18 backbone with 5-fold cross-validation on the AUC score.   
The ablation study about hard sample queue length is shown in the first two rows of Table. \ref{abla}. To balance the real-time and accuracy of queue updates, we choose 16, 32, and 64 as the length of queue $Q_h$. We find the length of 32 is appropriate, which achieves the highest AUC value. More importantly, all the AUC scores in this table significantly exceed the baseline by about 2\%, showing that discarding the samples with inconsistent labels is useful.
The ablation study about the certainty-aware queue is shown in the last two rows of Table.~\ref{abla}. By replacing the hyper-parameter $\alpha$ with the certainty-aware variable $\theta$, we achieve better performance while avoiding the laborious hyper-parameter selection.

Moreover, we further visualize the variation tendency of the $T_{ada}$ and the $\theta$ in Fig.~\ref{fig2}. The $T_{ada}$ is used to discover the hard sample and force the model to learn from easy to hard. The model certainty $\theta$ curve gradually increased to 1, which validates our assumption in Section 3.2.

\begin{table}[]
\centering
\caption{Comparison with the state of the art methods.}
\begin{tabular}{@{}ccccccc@{}}
\toprule
Learning strategy               & Backbone & Accuracy & Precision & Recall & F1-score & AUC   \\ \midrule
\multirow{3}{*}{Cross Entropy}  & resnet   & $70.75 _{\pm1.53}$   & $61.71 _{\pm1.64}$     & $68.05 _{\pm6.98}$  & $64.47 _{\pm2.50}$    & $77.31_{\pm1.04}$ \\
                                & densenet & $70.36_{\pm1.21}$  & $63.92_{\pm2.33}$     & $63.31_{\pm6.28}$  & $63.31_{\pm2.49}$    & $78.05_{\pm0.42}$ \\ 
                                & convnext & $72.29_{\pm1.66}$    & $62.94_{\pm2.62}$     & $71.19_{\pm8.77}$  & $66.36_{\pm2.84}$    & $78.38_{\pm1.04}$ \\
                                \midrule
\multirow{3}{*}{CL \cite{LyuT20curriloss}}     & resnet   & $70.54 _{\pm1.32}$   & $62.87 _{\pm3.62}$     & $65.85 _{\pm5.27}$  & $64.04 _{\pm1.70}$    & $76.98_{\pm1.16} $ \\
                                & densenet & $70.03_{\pm0.97}$    & $62.44_{\pm3.89}$     & $65.25_{\pm7.92}$  & $63.31_{\pm2.05}$    & $77.27_{\pm0.77}$ \\
                                & convnext & $72.14_{\pm1.74}$    & $60.21_{\pm2.67}$     & $\bm{75.76_{\pm3.65}}$  & $67.00_{\pm1.70}$    & $74.37_{\pm2.71}$ \\
                                 \midrule
\multirow{3}{*}{SL \cite{Castells20superloss}}     & resnet   & $71.26_{\pm1.82}$    & $\bm{64.97_{\pm1.73}}$     & $64.15_{\pm4.72}$  & $64.46 _{\pm2.67}$    &$ 78.50_{\pm1.24} $  \\
                                & densenet & $71.82_{\pm1.66}$    & $\bm{64.38_{\pm1.81}}$    & $66.86_{\pm5.84}$  & $65.42_{\pm2.49}$    & $78.98_{\pm0.29}$ \\ 
                                & convnext & $71.47_{\pm1.53}$    & $\bm{62.66_{\pm1.33}}$    & $68.39_{\pm3.46}$  & $65.36_{\pm1.99}$    & $77.87_{\pm1.07}$ \\
                                \midrule
\multirow{3}{*}{\textbf{ACL (Ours)}}  & resnet   & $\bm{73.28_{\pm0.53}}$    & $63.02 _{\pm1.86}$     & $\bm{73.64 _{\pm1.90}}$  & $\bm{67.87_{\pm0.40}}$    & $\bm{79.89_{\pm0.89}}$\\
                                & densenet & $\bm{72.92_{\pm1.26}}$ & $ 60.74_{\pm3.78}$    &$\bm{77.80_{\pm4.27}}$  & $\bm{67.99_{\pm0.88}}$   &$\bm{79.84_{\pm0.79}}$ \\ 
                                & convnext & $\bm{72.50_{\pm0.69}}$ & $61.19_{\pm1.68}$     & $74.75_{\pm3.65}$  & $\bm{67.21_{\pm0.91}}$ & $\bm{78.85_{\pm0.99}}$ \\
                                \bottomrule
\end{tabular}
\label{exp:sota}
\end{table}
\subsection{Comparison with the State-of-the-arts}
The comparison with the state-of-the-art methods are shown in Table. \ref{exp:sota}. These experiments are based on three widely used neural networks (ResNet18\cite{heresnet16}, DenseNet121\cite{huangdense17}, ConvNeXt-Tiny\cite{liu2022convnet}) power with the advanced curriculum learning based loss functions (Curriculum Loss (CL) \cite{LyuT20curriloss}, SuperLoss (SL) \cite{Castells20superloss})), and the proposed ACL. 
As shown in this table, the proposed method significantly exceeds the baseline by 3.5\% w.r.t averaging AUC score. It also outperforms other curriculum learning-based methods (i.e., SL and CL) that not change the backbone by more than 1\% AUC score on average, showing that properly discarding samples is quite effective.


\section{Conclusion}
We present an adaptive curriculum learning (ACL) framework for thyroid nodule diagnosis to resolve the {\em Inconsistent Label} problem, which adaptively discovers and discards the hard samples to constrain the neural network to learn from easy to hard. The proposed ACL could be easily embedded into existing neural networks to boost performance. Moreover, we contribute TNCD, a dataset that contains 3464 ultrasonic thyroid images with its cytological benign and malignant labels and pixel-level nodule masks. Extensive experiments have shown that the proposed framework outperforms state-of-the-art methods while unveiling that the "less is more" principle is practical.

\subsubsection{Acknowledgement.}
This work is supported in part by the Chinese Key-Area Research and Development Program of Guangdong Province (2020B0101350001), in part by the Guangdong Basic and Applied Basic Research Foundation \\(2020B1515020048), in part by the National Natural Science Foundation of China (61976250), in part by the Guangzhou Science and technology project (No.202102020633), and in part by the Guangdong Provincial Key Laboratory of Big Data Computing, The Chinese University of Hong Kong, Shenzhen.

\bibliographystyle{splncs04}
\bibliography{ref}
\end{document}